# Performance Evaluation of Image Enhancement Techniques on Transfer Learning for Touchless Fingerprint Recognition

S Sreehari[1], Dilavar P D[2], S M Anzar[3,a], Alavikunhu Panthakkan[4,b], and Saad Ali Amin[5] [1,3]Department of Electronics and Communication, TKM College of Engineering, Kollam, India [1,3]APJ Abdul Kalam Kerala Technological University, Thiruvanaanthapuram
[2]Center for Artificial Intelligence, TKM College of Engineering, Kollam, India [4,5]College of Engineering and IT, University of Dubai, Dubai, U.A.E
Corresponding Authors: [a]anzarsm@tkmce.ac.in, [b]apanthakkan@ud.ac.ae

Abstract—Fingerprint recognition remains one of the most reliable biometric technologies due to its high accuracy and uniqueness. Traditional systems rely on contact-based scanners, which are prone to issues such as image degradation from surface contamination and inconsistent user interaction. To address these limitations, contactless fingerprint recognition has emerged as a promising alternative, providing non-intrusive and hygienic authentication.

This study evaluates the impact of image enhancement techniques on the performance of pre-trained deep learning models using transfer learning for touchless fingerprint recognition. The IIT-Bombay Touchless and Touch-Based Fingerprint Database, containing data from 200 subjects, was employed to test the performance of deep learning architectures such as VGG-16, VGG-19, Inception-V3, and ResNet-50. Experimental results reveal that transfer learning methods with fingerprint image enhance-ment (indirect method) significantly outperform those without enhancement (direct method). Specifically, VGG-16 achieved an accuracy of 98% in training and 93% in testing when using the enhanced images, demonstrating superior performance compared to the direct method.

This paper provides a detailed comparison of the effectiveness of image enhancement in improving the accuracy of transfer learning models for touchless fingerprint recognition, offering key insights for developing more efficient biometric systems.

## I. INTRODUCTION

Fingerprint recognition has long been a reliable and widely used biometric method for personal identification and authen-tication [1]. Its popularity stems from the uniqueness and permanence of fingerprints, making them a highly depend-able characteristic for various applications, including secu-rity, forensics, healthcare, and device authentication [2], [3]. Traditionally, fingerprint identification systems have relied on contact-based scanners, which capture fingerprint images by direct physical contact between the user's finger and the sensor [4]. However, these systems are often susceptible to issues such as latent fingerprint residues, contamination, and image quality degradation due to varying pressure applied during contact. These limitations can negatively impact the accuracy and reliability of fingerprint recognition systems.

To overcome the challenges associated with contact-based systems, touchless fingerprint recognition has emerged as a promising alternative [5]. By utilizing digital cameras and advanced image processing techniques, touchless systems aim to capture fingerprint images without direct physical contact. This method not only enhances hygiene by eliminating contact but also mitigates problems such as smudging, distortion, and residual prints. However, the performance of touchless finger-print recognition systems is highly dependent on the quality of the captured images, making image enhancement techniques critical for achieving accurate and reliable recognition.

In recent years, deep learning, particularly Convolutional Neural Networks (CNNs), has revolutionized the field of computer vision and biometric recognition. Pre-trained CNN models, through the use of transfer learning, have shown re-markable success in image processing tasks, including finger-print recognition. Transfer learning allows models to leverage pre-trained knowledge from large datasets, reducing the need for extensive training on smaller biometric datasets. This study explores the effectiveness of image enhancement techniques combined with transfer learning models for touchless finger-print recognition. By evaluating pre-trained models such as VGG-16, VGG-19, Inception-V3, and ResNet-50, the study aims to determine the impact of image preprocessing on recognition performance and identify the optimal approach for achieving high accuracy in touchless fingerprint identification.

This paper presents a comprehensive evaluation of the impact of image enhancement techniques applied to touchless fingerprint images, comparing their impact on various deep learning models using transfer learning. Through experimental analysis using the IIT-Bombay Touchless and Touch-Based Fingerprint Database, this study demonstrates that fingerprint image enhancement significantly improves recognition accu-racy, with models utilizing enhanced images outperforming those without preprocessing. The findings provide valuable insights for developing more robust and reliable touchless fingerprint recognition systems, paving the way for further advancements in biometric security technologies.

## II. Literature Review

Touchless fingerprint recognition has emerged as a viable alternative to traditional contact-based methods, offering benefits in terms of hygiene, security, and user convenience. A central challenge in this field is improving image quality and recognition accuracy, where image enhancement techniques play a critical role. This study builds on prior research by examining the impact of image preprocessing on transfer learning models for touchless fingerprint recognition.

Preprocessing is widely recognized as crucial for improving fingerprint image quality. Sheregar et al. [4] emphasized its importance for enhancing feature extraction in touchless fingerprint recognition systems. Sagiroglu et al. [5] demonstrated that mobile touchless fingerprint systems with integrated enhancement pipelines significantly improved recognition performance, reinforcing the value of preprocessing. Similarly, Grosz et al. [6] addressed the challenge of matching contact-based and contactless fingerprints, highlighting the importance of consistent image quality through preprocessing to enhance matching accuracy.

Labati et al. [7] explored pore extraction using convolutional neural networks, underscoring the need for image enhancement to extract key features often lost in touchless fingerprints. This research aligns with our study's focus on evaluating enhancement techniques to optimize transfer learning models.

The effectiveness of transfer learning in image classification tasks has been validated in numerous studies. Tammina [8] demonstrated the strong performance of VGG-16 in classification tasks, making it a suitable model for touchless fingerprint recognition when paired with image enhancement. Lin and Kumar [9] also emphasized the role of preprocessing in aligning features between contactless and contact-based fingerprints, further supporting the necessity of image preprocessing for improving model performance.

Priesnitz et al. [10] reviewed touchless fingerprint recognition technologies and stressed the need for advanced image preprocessing to enhance system reliability. Their recommendations align with this study's objective of evaluating the impact of various image enhancement techniques. Furthermore, research by Choi et al. [11] and Galbally et al. [12] suggested that mosaicking techniques, when combined with preprocessing, could further improve fingerprint recognition systems.

In conclusion, the literature highlights the vital role of preprocessing and enhancement techniques in improving the accuracy of touchless fingerprint systems. This study builds on these findings by evaluating the impact of image enhancement techniques on the performance of transfer learning models, aiming to enhance the accuracy and reliability of touchless fingerprint recognition systems.

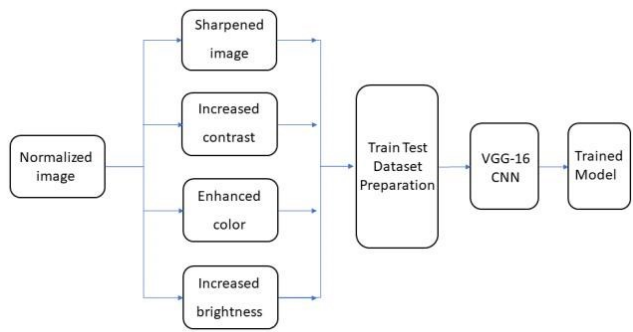

Fig. 1: Block diagram of Touchless fingerprint Identification without Preprocessing

## III. Background Work

In this approach, we expand the dataset for touchless fingerprint recognition using augmentation techniques to increase diversity and volume. Instead of traditional preprocessing, these augmentations prepare the data for training deep learning models. Geometrical augmentation is avoided to prevent distortion of fingerprint patterns. The methodology, as shown in Figure 1, generates varied samples to help models generalize better without altering the inherent fingerprint structure. Normalization ensures uniform pixel values, reducing computational complexity and improving feature interpretation (Figure 2(ii)).

Contrast adjustment enhances the distinction between bright and dark pixels, highlighting critical ridges and valleys (Figure 2(iii)). Color enhancement (Figure 2(iv)) amplifies dominant color channels, preserving important fingerprint details. Brightness is adjusted cautiously to avoid obscuring features (Figure 2(v)), while sharpening (Figure 2(vi)) fine-tunes edges and reduces noise. These augmentations enrich the dataset, allowing the deep learning models to generalize better and perform effectively in classification tasks, even without pre-processing.

## IV. Proposed Method

In this method, touchless fingerprint images undergo preprocessing to enhance critical features like ridges and patterns, improving their suitability for deep learning models. Once processed, the images are used for training and testing to optimize feature extraction and classification.

The process, shown in Figure 3.6, involves several key steps. First, normalization ensures uniform pixel values across the dataset, reducing computational complexity (Figure 3.7(i)). CLAHE (Contrast Limited Adaptive Histogram Equalization) then improves contrast by processing small image sections, avoiding over-amplification and enhancing visibility of key features (Figure 3.7(ii)).

Next, the SIFT (Scale-Invariant Feature Transform) algorithm detects and describes local features, generating high-dimensional descriptors for keypoints, critical for classification

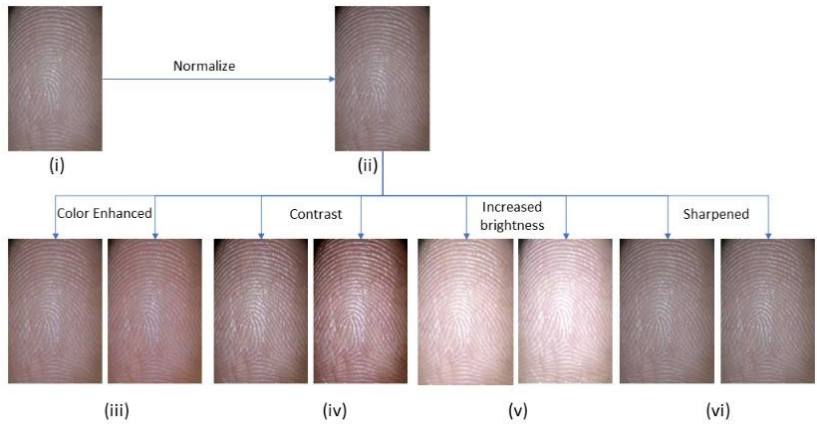

Fig. 2: (i) Input image, (ii) Normalized image, (iii) Color Enhanced image, (iv) Contrast increased image, (v) Increased brightness (vi) Sharpened image

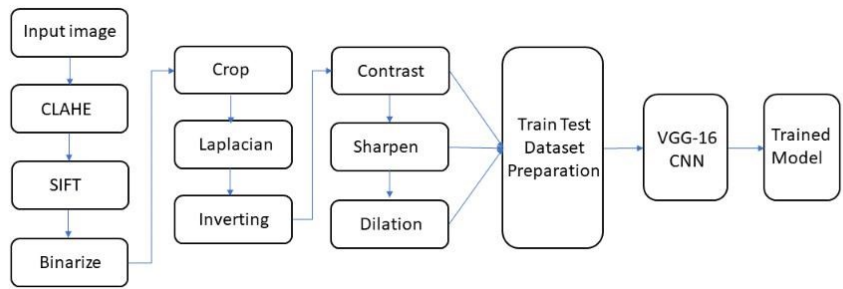

Fig. 3: Block diagram of Touchless fingerprint Identification with Preprocessing

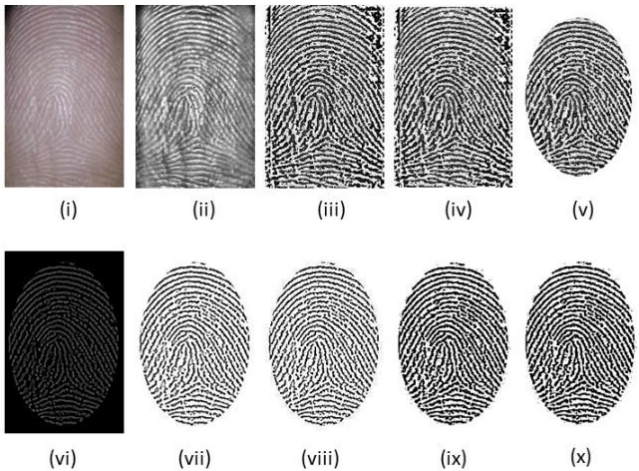

Fig. 4: (i) Normalization, (ii) CLAHE Enhancement Method, (iii) SIFT Algorithm, (iv) Threshold, (v) Cropping, (vi) Lapla-cian filter, (vii) Image inverting, (viii) Sharpening, (ix) Con-trast, (x) Dilation

(Figure 3.7(iii)). Thresholding converts grayscale images to binary, clearly separating fingerprint patterns from the back-ground (Figure 3.7(iv)).

Cropping removes irrelevant background noise, focusing on the fingerprint area (Figure 3.7(v)), while the Laplacian filter sharpens edges and highlights ridges and valleys (Figure 3.7(vi)). Image inversion adjusts the binary values to ensure correct representation (Figure 3.7(vii)). Contrast and sharpen-ing further clarify the image (Figure 3.7(viii-ix)), and dilation strengthens boundary lines by adding pixels to edges, closing gaps in the fingerprint pattern (Figure 3.7(x)).

These preprocessing steps optimize the fingerprint images for deep learning models, enhancing feature extraction and improving classification accuracy.

## V. RESULTS AND DISCUSSION

### A. Datasets

The dataset used in this study is the IIT-Bombay Touch-less and Touch-Based Fingerprint Database, containing 800 touchless and 800 touch-based fingerprint images from 200 subjects, with four samples per subject. The touchless images (170 x 260 pixels) were captured using a Lenovo Vibe K5 Plus

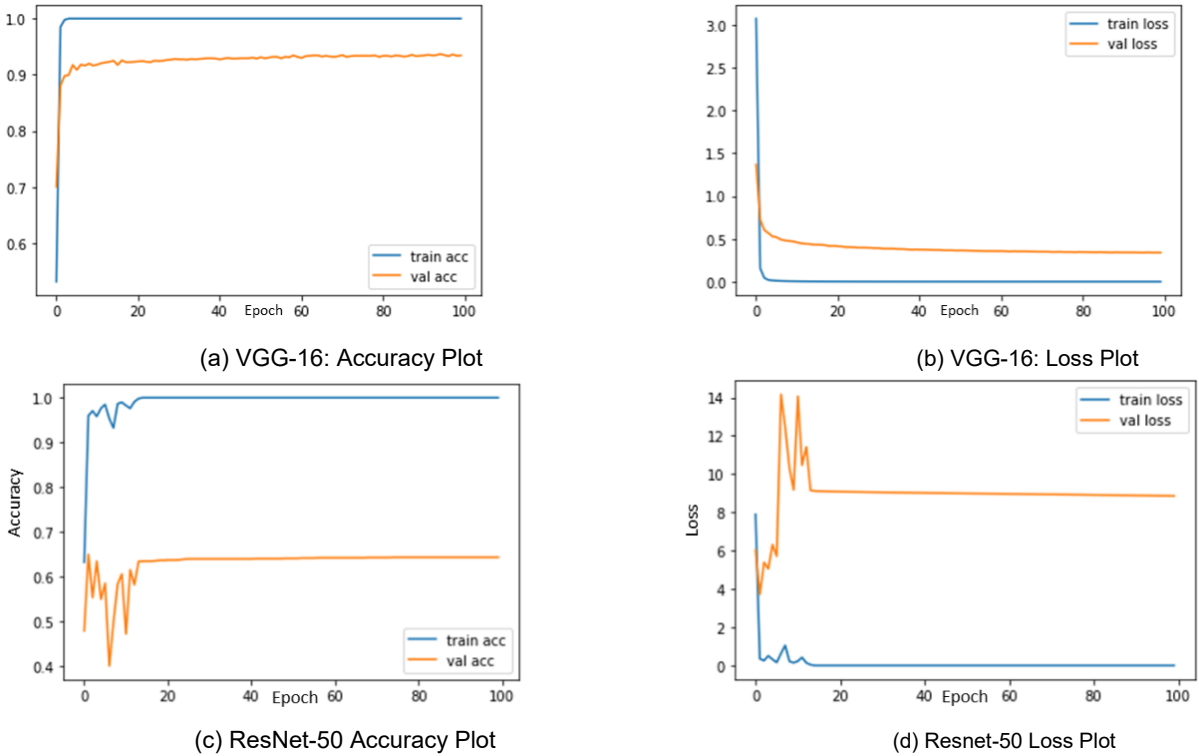

(a) VGG-16: Accuracy Plot

(b) VGG-16: Loss Plot

(c) ResNet-50 Accuracy Plot

(d) Resnet-50 Loss Plot

Fig. 5: Accuracy and Loss Plots Vs Epoch of Transfer Learning Models without Preprocessing

TABLE I: Performance of Transfer Learning Models for Touchless Fingerprints without Preprocessing

| S.No | Model | Accuracy | Loss | Precision | Recall | F1-score |
|------|-------|----------|------|-----------|--------|----------|
| 1 | VGG-16 | .93 | 33.86 | 0.95 | 0.93 | 0.94 |
| 2 | VGG-19 | 0.92 | 0.38 | 0.95 | 0.90 | 0.93 |
| 4 | ResNet-50 | 0.64 | 0.88 | 0.64 | 0.64 | 0.64 |
| 3 | Inception-V3 | 0.17 | 1.85 | 0.19 | 0.14 | 0.16 |

TABLE II: Performance Evaluation of Transfer Learning Models for Touchless Fingerprints with Preprocessing

| S.No | Model | Accuracy | Loss | Precision | Recall | F1-score |
|------|-------|----------|------|-----------|--------|----------|
| 1 | VGG-16 | 0.98 | 0.20 | 0.99 | 0.93 | 0.98 |
| 2 | VGG-19 | 0.97 | 0.24 | 0.99 | 0.93 | 0.97 |
| 3 | ResNet-50 | 0.76 | 1.03 | 0.84 | 0.69 | 0.76 |
| 4 | Inception-V3 | 0.64 | 1.16 | 0.64 | 0.63 | 0.63 |

smartphone, while the touch-based images (260 x 330 pixels) were slightly larger.

The dataset was structured for training and testing deep learning models, with each subject's images stored in separate folders, resulting in 200 distinct classes. Two methods were used: one without preprocessing and one with preprocess-ing. In the method without preprocessing, the dataset was augmented, resulting in 3,200 images for training and 1,600 for testing, across 200 classes. Augmentations were applied directly to the raw images to assess model performance without prior enhancement. In the preprocessing method, images underwent enhancement steps to improve quality. After augmentation and preprocessing, 1,800 images were used for training and 600 for testing. This approach aimed to enhance key features, allowing the models to extract more robust features for improved classification, enabling a comparative analysis of preprocessing's impact on performance.

B. Experimental Set-up

The program for this study was implemented in Python, leveraging its robust libraries for machine learning and deep learning. The hardware setup included an Intel i5 10th generation processor with 8GB RAM and an NVIDIA GeForce GTX 1650 GPU to accelerate training. Jupyter Notebook was used for development and execution,

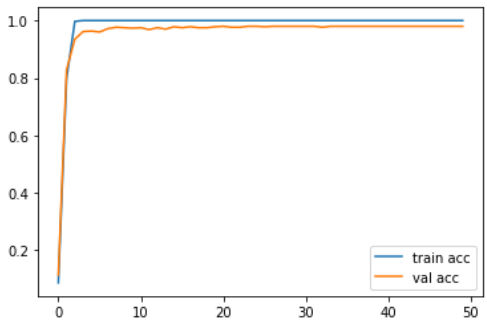

(a) VGG-16: Accuracy Plot

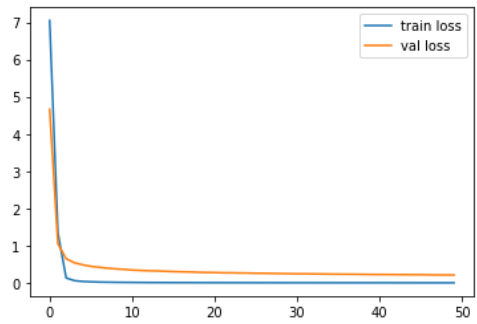

(b) VGG-16: Loss Plot

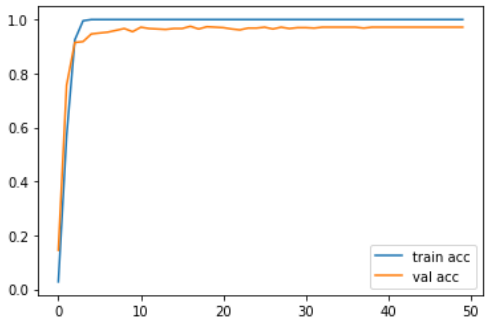

(c) VGG-19 Accuracy Plot

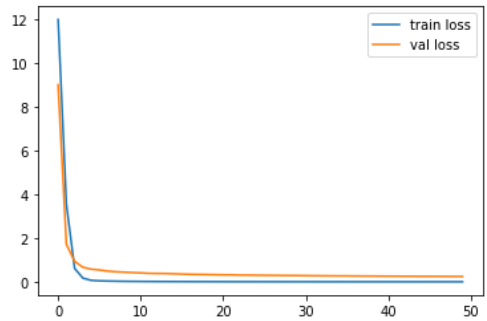

(d) VGG-19 Loss Plot

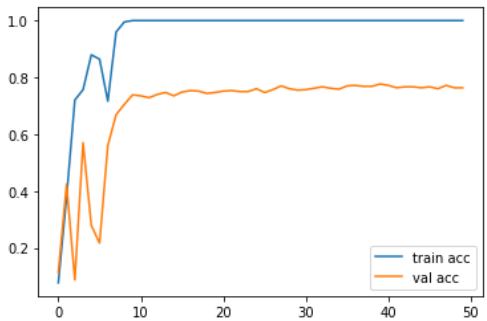

(e) ResNet-50 Accuracy Plot

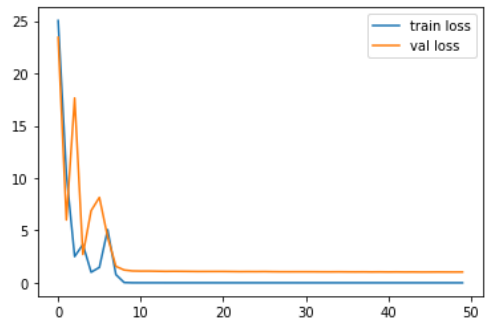

(f) Resnet-50 Loss Plot

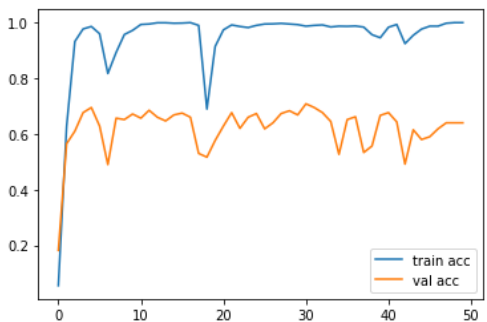

(g) Inception-v3 Accuracy Plot

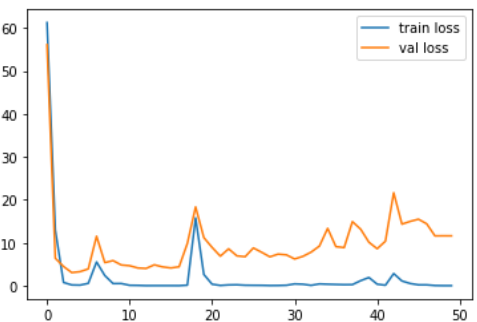

(h) Inception-v3 Loss Plot

Fig. 6: Accuracy and Loss Plots Vs Epoch of Transfer Learning Models with Preprocessing

facilitating interactive coding and visualization. These configurations ensured smooth and efficient model training. Notebook presets and hyper-parameter setups

- Architecture = VGG-16, VGG-19, Inception-V3 and ResNet-50
- Loss = categorical crossentropy
- Optimizer = Nadam

- Target size = 224 x 224 pixel
- Batch size = 32
- Class mode = categorical
- Epochs = 100

C. Performance Comparison With and Without Preprocessing

In comparing the performance of transfer learning models for touchless fingerprint recognition with and without preprocessing, it is evident that preprocessing significantly improves the outcomes across all models. Without preprocessing, VGG-16 and VGG-19 lead the performance, with VGG-16 achieving 93% accuracy and an F1-score of 0.94, while VGG-19 follows closely with 92% accuracy and an F1-score of 0.93. However, ResNet-50 and Inception-V3 perform poorly, with Inception-V3 displaying the lowest accuracy at just 17% and a high loss of 1.85, highlighting its struggle in this scenario.

In contrast, the models exhibit notable improvements with preprocessing. VGG-16 reaches 98% accuracy and an F1-score of 0.98, while VGG-19 closely matches with 97% accuracy and an F1-score of 0.97. ResNet-50 also shows substantial improvement, increasing its accuracy to 76% and F1-score to 0.76. Even Inception-V3 benefits from preprocessing, achiev-ing 64% accuracy, though it still underperforms compared to the other models.

The accuracy and loss plots further illustrate that both VGG models converge smoothly in both scenarios, but show better stability and reduced loss with preprocessing. ResNet-50, while more variable, benefits from preprocessing with better stabilization. Inception-V3, however, remains inconsistent and exhibits poor learning, even with preprocessing.

In summary, preprocessing significantly boosts model performance, particularly for VGG-16 and VGG-19, while also benefiting ResNet-50. Inception-V3 continues to perform poorly, but still shows improvement with preprocessing.

## VI. Conclusion and Future Work

This study evaluated the performance of image enhancement techniques on transfer learning models for touchless fingerprint recognition using VGG-16, VGG-19, Inception-V3, and ResNet-50. Two approaches were employed: one with preprocessed images and the other without preprocess-ing. Among the models, VGG-16 achieved the highest ac-curacy, with 98% for preprocessed images and 93% without preprocessing, highlighting the significant impact of image preprocessing on model performance. While the other models showed similar trends, none outperformed VGG-16. These results confirm that deep learning, particularly when combined with image enhancement techniques, is highly effective for touchless fingerprint identification.

Future research should focus on integrating additional depth features, such as 3D fingerprint capturing and mosaicking, to further improve performance. Moreover, the inclusion of novel biometric traits, such as sweat pore pattern analysis and blood flow detection, could further enhance system accuracy and security. These advancements will elevate touchless fingerprint recognition to a more secure and reliable method of biometric authentication in the rapidly evolving field of cybersecurity.